# R4: rapid reproducible robotics research open hardware control system

Chris Waltham
Andy Perrett
Rakshit Soni
Charles Fox
*School of Engineering and Physical Science, University of Lincoln, UK.*

## Abstract

A key component of any robot is the interface between robotics middleware software and physical motors. New robots often use arbitrary, messy mixtures of closed and open motor drivers and error-prone physical mountings, wiring, and connectors to interface them. There is a need for a standardizing OSH component to abstract this complexity, as Arduino did for interfacing to smaller components. We present a OSH printed circuit board to solve this problem once and for all. On the high-level side, it interfaces to Arduino Giga – acting as an unusually large and robust shield – and thus to existing open source software stacks. A ROS2 interface is provided. On the lower-level side, it interfaces to existing emerging standard open hardware including OSH motor drivers and relays, which can already be used to drive fully open hardware wheeled and arm robots. This enables the creation of a family of standardized, fully open hardware, fully reproducible, research platforms.

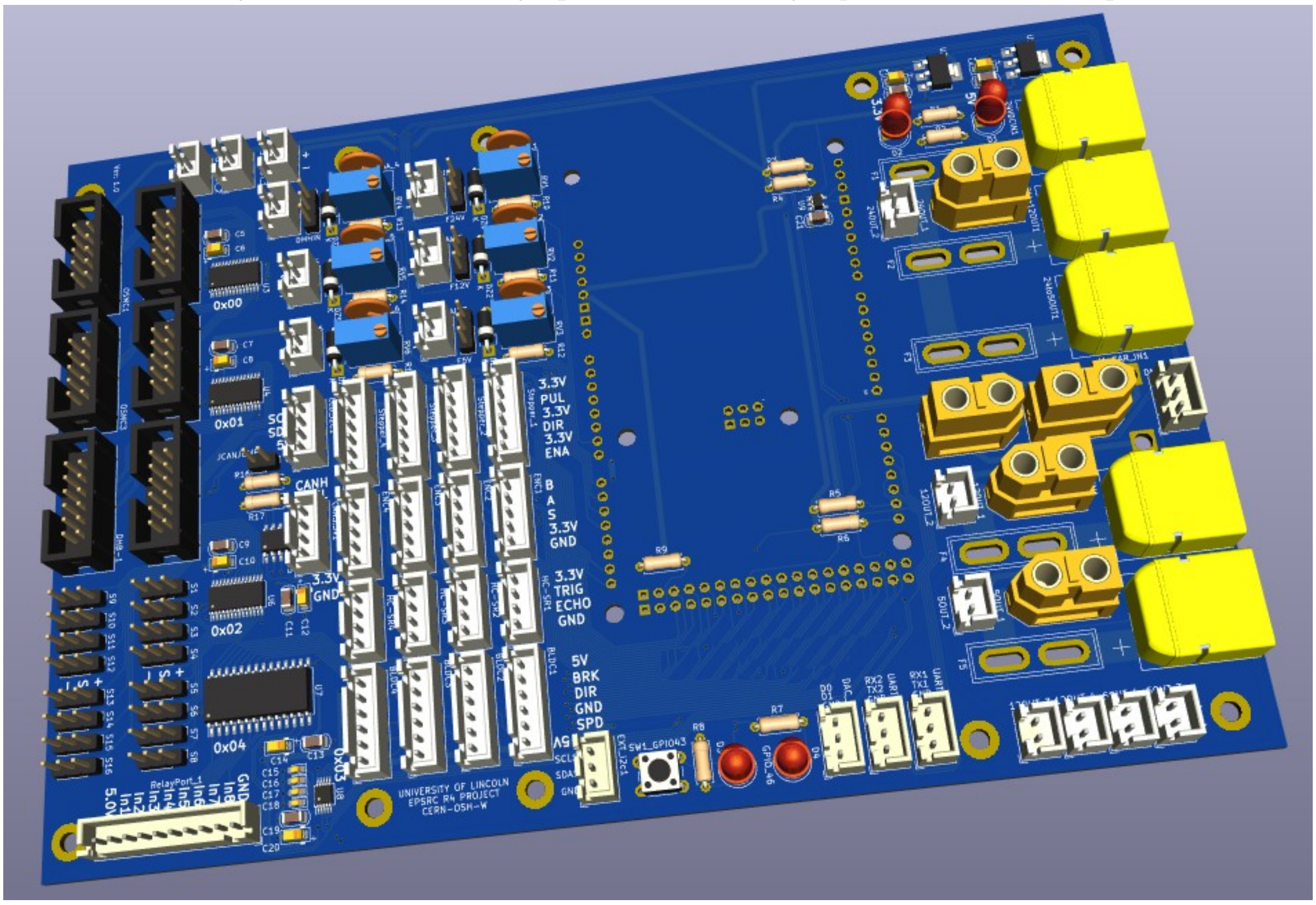

Fig 1. R4 board KiCAD model

## Metadata Overview

Main design files: https://doi.org/10.5281/zenodo.15191374
Target group: secondary school students, undergraduate students, scientists in robotics
Skills required: PCB manufacture and PCBA assembly, or outsourcing. Arduino firmware configure and install.
Build cost: 300 USD
Replication: Five R4 boards have been successfully built and tested, by different researchers in our lab following the build instructions, using pcbway PCB manufacture and manual placing and reflow over for PCBA assembly.

## Introduction

Robotics research still struggles with reproducibility. Open source robotics middleware such as ROS2 [1] and YARP [2] enable reuse of software, but not hardware. Researchers waste time porting systems between hardware platforms to reproduce research between labs only approximately, and it is rare to be able to reproduce specific published experimental results [3]. Researchers in developing countries often cannot afford the proprietary robots used by labs with more resources. If a published robotics system is dependent on any component that is only available from a single supplier, then all work building on it is at risk if that supplier vanishes, de-lists or changes the product. Combined open source software and open source hardware stacks allow both researchers and industry to download, build, exactly replicate, deploy or extend the published work which they read about.

Open Source Hardware (OSH, [4]) is hardware whose designs and build instructions are public, easy, and low-cost so that anyone is free to build and modify them, enabling large community collaborations. Recent `shallow' definitions of OSH such as the 2020 CERN-OHL licences [5] do not require designs to be made up of entirely OSH sub-components, rather they allow for closed source sub-components if they are available on the open market. However, “deep” definitions such as those aspired to by Open Source Ecology [6] further restrict subcomponents, recursively, to be all OSH, so that entire designs are open down to the level of ISO standard nuts, bolts, resistors and transistors. “Deeper OSH” has been proposed [7] as the process of successively replacing lower level sub-components of shallow OSH designs with OSH alternatives, working towards deep OSH. To enable this process, there is a need for standard, widely reusable sub-components to be created as OSH.

A key component of any robot is the interface between robotics middleware software and physical motors. New robots often use arbitrary, messy mixtures of closed and open motor drivers and error-prone physical mountings, wiring, and connectors to interface them. There is a need for a standardizing OSH component to abstract this complexity, as Arduino did for interfacing to smaller components. We present a OSH printed circuit board, R4, to solve this problem once and for all. On the high-level side, it interfaces to Arduino Giga – as an unusually large and robust shield – and thus to existing open source software stacks. We provide a socket-based protocol and include a ROS2 wrapper. On the lower-level side, it interfaces to existing emerging standard open hardware including OSH motor drivers and relays, which can already be used to drive fully open hardware wheeled and arm robots. This enables the creation of a family of standardized, fully open hardware, fully reproducible, research platforms. Like Arduino, the new standard may also be taken up by industry to speed up their development, especially via knowledge transfer from research to commercial systems.

The R4 board is designed for use in what we choose to define as *medium sized robots*: being robots capable of moving one person or a load which could be carried by one person, while not being capable of causing major injury in a collision with a person. This includes last mile delivery vehicles, precision (but not bulk operation) agricultural robots, and personal human micromobility vehicles. It could also include industrial robot arms and consumer mechatronics devices such as vending machines, door and gate automations, and animatronics models.

R4 is designed with a single, supplied, firmware, which does not need to be modified by users. This removes the need for users to write embedded code for their own devices. Rather, control of new devices can be described at higher levels including inside ROS2 while R4 handles translation to the lower levels.

## Overall Implementation and design

R4 features are designed to accommodate many typical use cases for medium-sized mobile robots, robot arms, and conveyor style systems, including differential and Ackerman steering, open and closed loop control, and low-level safety:

**Dead Man's handle:** This electrical level safety interlock consists of a handheld button which controls one or more relays. The switch is wired in such a way as to provide a method of disconnecting driver motors should a safety issue arise via a relay (this relay can control other safety measures if required). The coil DMH circuit is also routed through another relay controlled by R4's firmware allowing software at the firmware or higher levels to act as an emergency stop. This ensures users must be present and in charge, and it provides a prompt means that can be controlled manually or from software to prevent a robot's movement where a potentially dangerous situation may occur, such as an imminent collision.

**Form factor:** The board dimensions are 197x136mm.

**Robust connectors:** XT-60 (power) and JST-XH (signals) connectors are provided. Practical vehicles in less-than-ideal environments often vibrate and dislodge weak connections. An Arduino Giga R1 is connected to the PCB via its headers, robust connectors throughout enable easy reuse of the PCB between different robots.

**Power Supplies:** The board is powered directly from 24V or 12V batteries or external power supply. It provides three high current outputs and one low current output. The high current outputs may be sent to external buck converters which return into the board to provide high current 12V and 5V supplies, which are then distributed to various interfaces. These supplies are each fused on the board and high current tracing has been designed, these high current traces define R4's maximum current ratings specifications. High reliability automotive grade ATC fuses are used. There are three low power and one high power 12 V and 5 V outputs, eight in total placed such that external boards can be powered. The board uses these externally derived 12 V and 5 V supplies for its interfaces where required. Separate 5 V and 3.3 V onboard regulators are supplied by the main input voltage connector and are used to power the micro controller and board ICs. A precision 3.3 V reference supply is also provided, which connects to the Giga's Aref pin improving absolute voltage measurements via the micro controllers ADC pins. The board is designed for a temperature rise of 28 ℃ for a maximum current of 20 A at 24 V as expected when used in typical medium-sized robots. Higher power actuators can be controlled `off board'.

**Analog Input channels:** There are 6 Analog Input channels connected to the ADC pins of the Giga. These channels have a voltage divider circuit, which can be configured using the preset to divide the incoming voltage. This allows a maximum input of 50V at each analog input channel. Zener clamps protect against over voltage and are fused to prevent damage to the Giga's ADC pins. Four of the 6 analog input channels have an optionally connectable jumper which measures various on-board voltage signals on the power supplies, two further channels have no jumpers. All are socketed to provide easy robust connection and monitoring of voltages.

**8 channel relay board interface:** This interface is used to control and switch an 8 channel relay board and control each relay signal separately, this uses one port of a 16 channel GPIO expander IC. These relay signals can be used to switch 10 A loads at up to 24 V.

**Stepper motor channels:** The board has 4 stepper motor channels which can connect to stepper motor drivers such as Microstep M542. Stepper speed and direction is controlled by timed signals generated by library code running on the Giga and can be set and monitored via the UDP link.

**Servo channels:** The board has the capability of interfacing 16 PWM 5V servo motors directly.

**OSMC H-bridge channels:** The board provides interfacing channels for Open Source Motor Control (OSMC, [8]) drivers. This allows usability of a wide range of motors 13 V-50 V and up to 160 A max current, although current as high as 160 A is stated by the OSMC documentation this is not recommended without further testing and validation.

**DHB H-bridge interfaces:** The reference used to design this interface is a DHB-12, which is a dual motor driver H-bridge that can control two 5 V-15 V motors at 30 A current. The board has two interfaces of this type allowing four more motors to be controlled separately for low to medium power applications.

**Expansion ports:** additional ports are provided which can be used to add new hardware if their Arduino drivers are wrapped as R4 software modules. This could include for example BLDC brushless motor drivers, motor encoders and ultrasonic sensors. It is also possible to reuse the other ports for new hardware by swapping their software drivers.

Fully automated or manual PCBA build are both enabled by the design. Its CAD format and selected components are designed to be compatible with major pick-and-place manufacturers. This enables new users to order an assembled PCB with just a few clicks. Alternatively, manual placing can be performed. The board is designed so that some features and components can be mounted optionally as per the use case to lower build cost.

The hardware and software of R4 are designed to reduce or remove the need for most end-users to write embedded code, while still allowing them to use R4 to control many different actuators and sensors. R4's firmware allows the user to control selected standard hardware via its open source UDP WiFi protocol, the user simply issues a command via the link to set the state of the hardware. For example, the PWM value of a servo can be directly set by sending a control character, a channel number and the value for the servo position. In a similar way the output direction of a H-Bridge channel can be set along with its duty cycle. R4's communications protocol also allows a user to set a stepper motors position by sending a command containing a channel and a requested position. Likewise, it is possible for a set of commonly used sensors to be connected. R4 can initialize sensors and report sensor data across the link in a time-controlled manor, i.e. sensor samples may be collated and sent via a datagram to the host at a defined sample period.

Each hardware component is defined by a software device driver written as a file of Arduino C. We provide a collection of such device drivers for a selection of commonly used standard components, including OSH motor drivers, servos, steppers, relays, and analog I/O. A driver defines input and output items in the string UDP protocol, and functions for initialisation, step, and emergency shutdown. In most cases they can be written by wrapping existing their Arduino libraries in these functions. Drivers are then selected and assigned to ports on the R4 board in the board's main Arduino function. In the default configuration, drivers are assigned to ports whose names (printed on the PCB) match their default intended purpose, such as OSMC and STEPPER, though it is possible to reuse the ports for any driver by changing these assignments.

At the ROS2 level, each hardware device is represented by a ROS2 Node, and the R4 board is represented by an additional ROS2 Node. The board's Node reads requests to send individual string commands to the board, as individual named ROS2 messages concerning different actuators, and concatenates them into a single string and passes this string to the board over UDP. It then reads strings back from the board and splits them into individual data reports concerning different sensors as separate named ROS2 messages. Device nodes then read and write these messages and convert them into physically meaningful values and units. This allows user code to interact with devices by passing physically meaningful ROS2 messages.

For example, a motor driver's software driver might define hexadecimal values corresponding to desired physical voltage levels. These are used in the UDP protocol, but the device ROS2 translates them into continuous voltages. An odometry sensor might similar speak hex over UDP but be converted by its ROS2 node into meters per second values. The user could then for example write a closed loop controller as a higher ROS2 node which outputs voltages and reads meters per second encoder feedback. This style of closed loop creation is appropriate in some cases, being easy to write at the ROS2 level. In other cases – such as brushless motor control – lower latency may be needed, in which case closed loops can we implemented inside the device driver software and the USP and ROS2 messages used only to specify the desired output level.

The rest of the board and communications protocol abstracts typical robotics housekeeping and safety tasks. This includes a heartbeat system which checks for continual ROS2 communications and safety shuts down each device if it is not present. It also includes safety fuse, and a dead man’s handle connected directly to a physical relay for instant power shutoff in emergencies. Other housekeeping includes accepting either 12 V or 24 V input – common robotics voltages – and providing 12 V, 5 V and 3.3 V supplies used by many devices. The software does not use a heavyweight RTOS but instead includes its own lightweight timer code to periodically call out to the set of device drivers.

The communication protocol adds timestamps and CRC error correction and lightweight UDP methods for resending lost or corrupted messages rather than using heavier weight TCPIP.

Where high frequency closed loops are required, such as for brushless motor control, the motor driver hardware connected to R4 typically executes this, three phase motor coils switching often occurs at higher frequencies than achievable across a UDP link.

R4 communicates with an external host using a human-readable ASCII line-based protocol. The protocol consists of carriage return terminated delimited strings (datagrams). The datagrams can be communicated either over a serial port or a UDP socket. R4 handles sampling hardware status, timing of datagrams, heartbeats, and asynchronous reception of command packets, this functionality is driven by Giga R1 hardware interrupts to maintain controlled timed sampling of the current hardware status.

Datagrams are streamed out continuously by R4 during normal operation, reporting the state of the hardware at timed intervals, and consisting of human-readable strings. R4 also reads human-readable command strings sent to it, and also reads heartbeat strings, both sent over UDP. If timely heartbeats are not received continuously then R4 initiates controlled shutdown of each of its device drivers. Full details of the protocol format are provided in the User Guide.

# Quality control

R4 is intended for use in medium sized robots such as those which can carry a person or person-sized load, so safety is of critical importance. The design features a dead man’s handle to cut off all power instantly in emergencies. It further includes fuses on each of the power supplies. A heartbeat system is used to ensure that the command connection is working at all times, with driver-specific safe shutdowns otherwise. Acceptance and integration tests are integrated in the build instructions.

# Application

## OpenScout Python UDP control

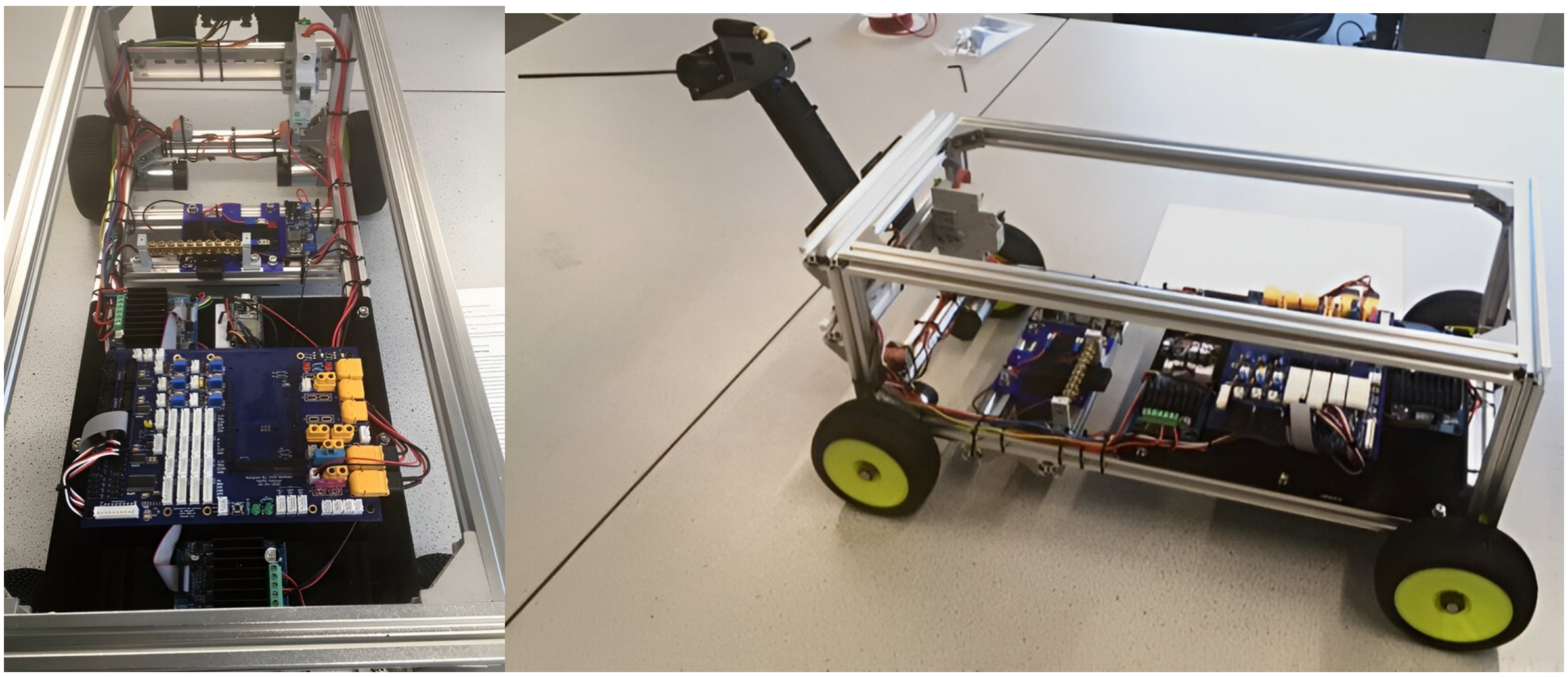

Fig 2. OpenScout R4 variant, from top and side.

R4 has been used in a student project to control four-wheel motors using two dual DHB12 H-bridges and a three-servo arm, replacing the electronics of the OSH OpenScout [9] mobile robot (fig. 2). A Python script communicates with R4 using R4 protocol over UDP via a Python socket library. Code is available in examplesUses/OpenScoutR4.

## PS5 control of an OSMC driver

A simple example was tested to use a PS2 controller for closed loop control of an encoder linear actuator, driven by an OSMC and using feedback from the inbuilt actuator encoder. Code is provided in examplesUses/PS2controller_OSMCcontrol.

## OpenPodcar ROS2 control

R4 has been used to replace the previous ad-hoc electronics in the OSH OpenPodCar [10] vehicle, as shown in fig. 3. R4 is here connected to an OSMC driver which controls the rear motor; and to a DHB12 which drives a linear actuator to turn the steering column with  an R4 ADC channel is used to sense the position of the linear actuator. The DMH system is also incorporated providing safety cut-off functionality preventing the main motor driving the vehicle if the DMH is not pressed. The vehicle has performed several demonstrations under manual control using an XBox controller connected to a laptop running ROS2. rvis2 has also been used to provide waypoint command messages, so that ROS2 nodes then generate R4 commands for movements to those way points within the map. The vehicle has been driven around 1 km in each of these manual and autonomous modes without failure. Code is in examples Uses/OpenPodcar_v2.

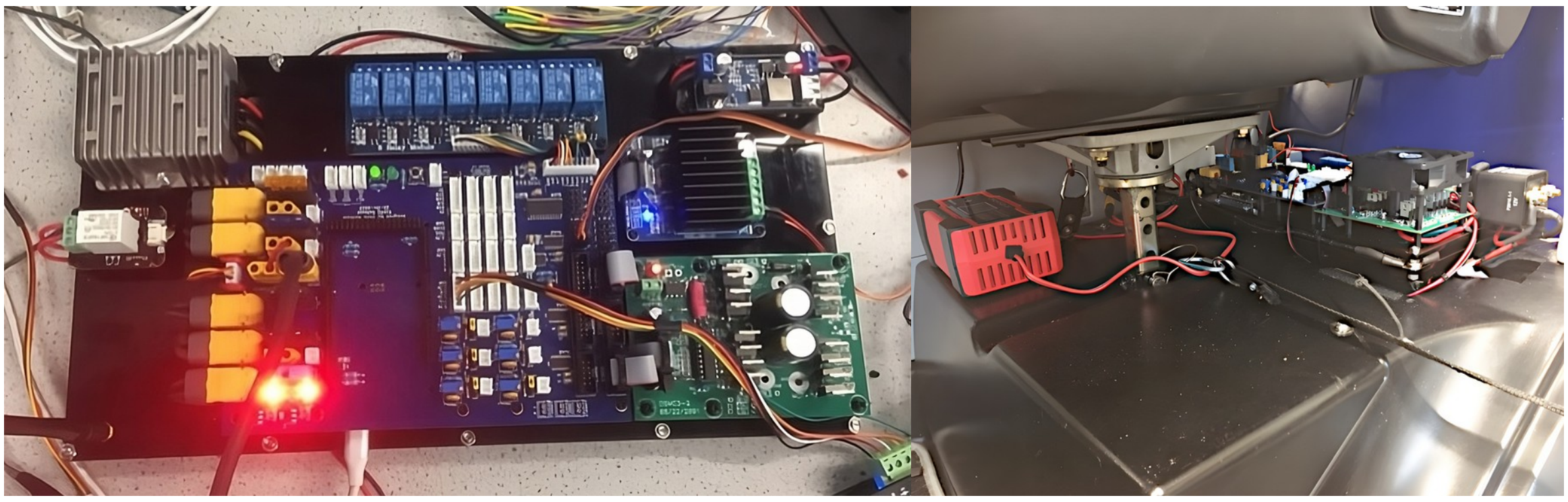

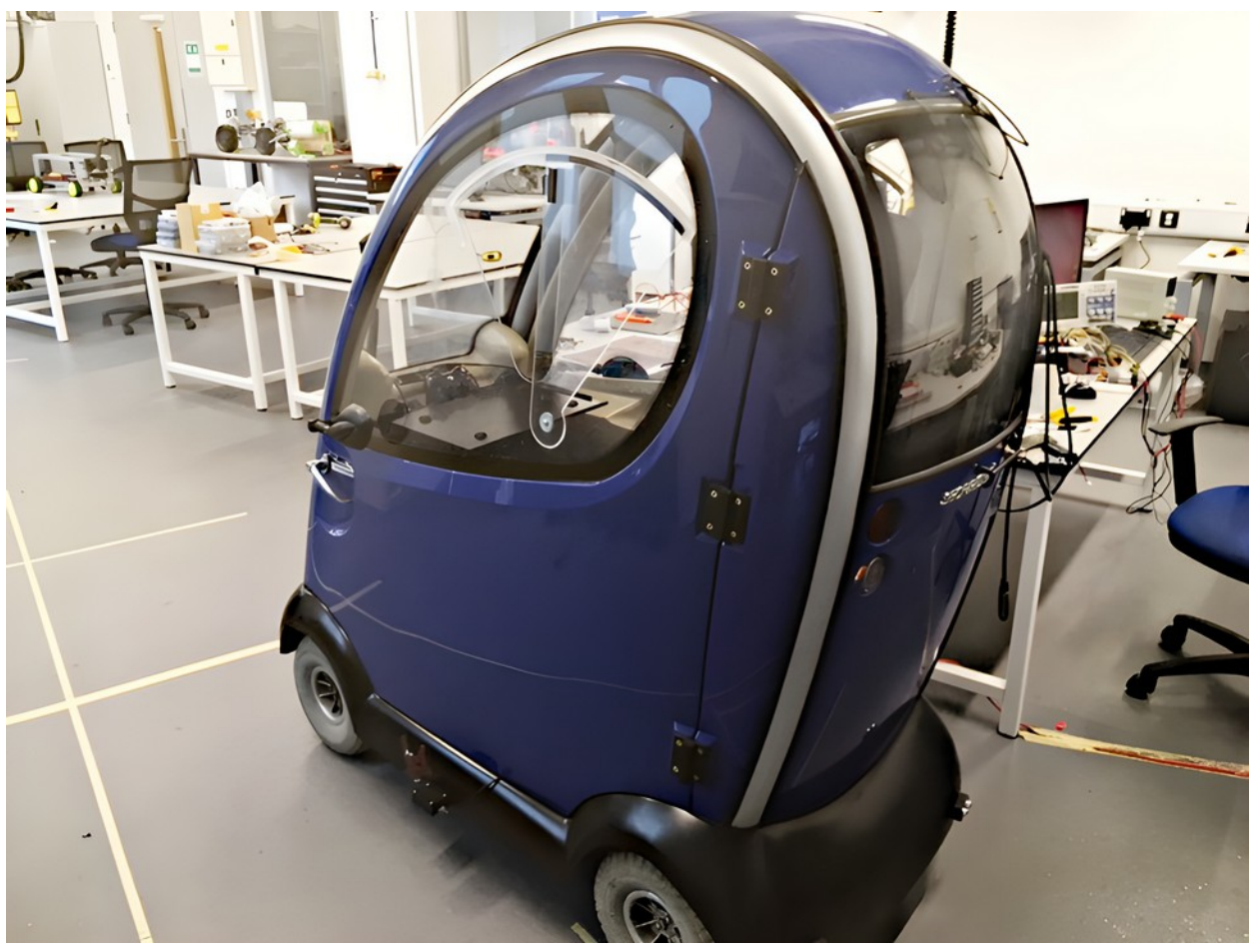

Fig 3. OpenPodcar R4 variant.
(a) R4 PCB and external drivers mounted on perspex; (b) Perspex board installed in vehicle under seat; (c) exterior.

# Build Details

## Availability of materials and methods

R4 is a printed circuit board assembly (PCBA) which uses a printed circuit board (PCB) with typical surface mounted and other electronics components available from typical electronics suppliers and PCBA automated manufacturers.

## Ease of build / Design decision

The PCBA can be sent to an external manufacturer for automated manufacture or assembled by hand. Typical components are deliberately used to enable automated manufacture by typical external PCBA manufacturers.

Manual assembly requires skilled technician skill level. The PCB can be sent out for manufacture or could be made locally where skills and tools are available. Assembling the PCBA then includes using a solder mask, placing surface mount technology (SMT) components by hand (with tweezers and a magnifying glass) and using a reflow oven to bake them on.

The schematic is shown in Fig. 4. It is not necessary to refer to the schematic during a normal build, but it may be useful for identifying problems with builds not caught by the integrated tests, or for modification of the design.

The BOM is summarised in Fig. 5. The full BOM spreadsheet in the repository also includes links to possible suppliers. The repository also includes copies of datasheet for all BOM components. Acceptance and integration tests are embedded throughout the build instructions.  End-user validation tests (i.e. user acceptance tests of the build) are provided in the user instructions.

## Operating software and peripherals

Arduino Giga is used with open source software firmware provided in the release. Details of compiling and uploading the firmware are provided in the user instructions. The firmware communicates with an external computing using a text protocol over UDP. The user is assumed to have basic Arduino skills such as flashing LEDs.

A ROS2 interface is also included, which wraps the underlying text protocol. ROS2 setup and use is described in the user guide. The user is assumed to have basic ROS2 skills such as completing the standard ROS2 tutorials. The ROS2 nodes and messages are shown in Fig. 6.

## Dependencies

R4 is intended to be extensible with drivers for any components, but is supplied with drivers for the open source hardware OSMC and DHB12 motor drivers; and standard stepper motors and relays.

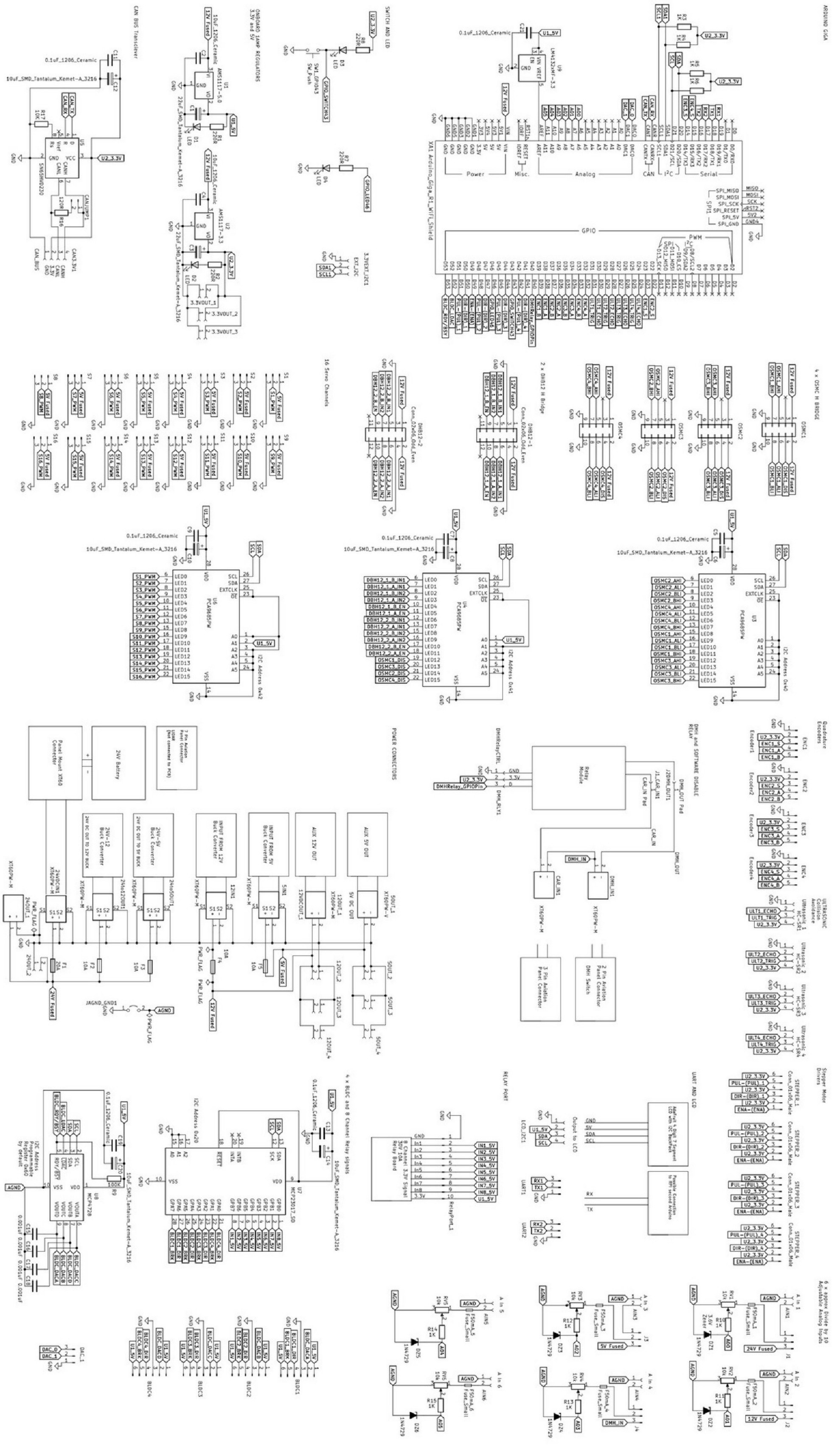

Fig 4. R4 schematic

| Footprint | Qty | Manufacturer | USD/unit |
|---|---|---|---|
| PCM_arduino-library:Arduino_Giga_R1_WiFi_Shield | 1 | Arduino | 79.58 |
| Capacitor_Tantalum_SMD_AKL:CP_EIA-3216-18_Kemet-A | 2 | KEMET | 0.76 |
| Capacitor_SMD:C_0603_1608Metric | 4 | KYOCERA AVX | 2.03 |
| Capacitor_SMD_AKL:C_1206_3216Metric | 2 | Murata | 4.77 |
| Capacitor_SMD_AKL:C_1206_3216Metric | 7 | Murata | 6.83 |
| Capacitor_Tantalum_SMD_AKL:CP_EIA-3216-18_Kemet-A | 6 | KEMET | 0.50 |
| Resistor_THT:R_Axial_DIN0207_L6.3mm_D2.5mm_P7.62mm_Horizontal | 4 | RS-Pro | 1.90 |
| Resistor_THT:R_Axial_DIN0207_L6.3mm_D2.5mm_P7.62mm_Horizontal | 1 | RS-Pro | 1.90 |
| Resistor_THT:R_Axial_DIN0207_L6.3mm_D2.5mm_P7.62mm_Horizontal | 10 | RS-Pro | 1.43 |
| Resistor_THT:R_Axial_DIN0207_L6.3mm_D2.5mm_P7.62mm_Horizontal | 2 | RS-Pro | 1.90 |
| Potentiometer_THT:Potentiometer_Bourns_3296W_Vertical | 6 | Bourns | 2.66 |
| Package_TO_SOT_SMD:SOT-223-3_TabPin2 | 1 | Advanced Monolithic Systems | 1.82 |
| Package_TO_SOT_SMD:SOT-223-3_TabPin2 | 1 | Advanced Monolithic Systems | 0.85 |
| Package_SO:TSSOP-28_4.4x9.7mm_P0.65mm | 3 | NXP | 3.58 |
| Package_SO:SOIC-8_3.9x4.9mm_P1.27mm | 1 | Texas Instruments | 2.61 |
| Package_SO:SOIC-28W_7.5x17.9mm_P1.27mm | 1 | Microchip | 2.08 |
| Package_SO:MSOP-10_3x3mm_P0.5mm | 1 | Microchip | 0.00 |
| Package_TO_SOT_SMD:SOT-23-5 | 1 | Texas Instruments | 3.08 |
| Fuse:Fuse_Blade_ATO_directSolder | 4 | Amazon | 8.09 |
| Fuse_US_AKL_Double:Fuse_BelFuse_0ZRE0005FF_L8.3mm_W3.8mm | 6 | Bel | 0.43 |
| LED_THT:LED_D5.0mm RED | 2 | Kingbright | 2.04 |
| LED_THT:LED_D5.0mm GREEN | 2 | Kingbright | 0.78 |
| Diode_THT_AKL:D_DO-41_SOD81_P7.62mm_Horizontal_Zener | 6 | onsemi | 0.31 |
| Button_Switch_THT:SW_PUSH_6mm | 1 | | 0.00 |
| Connector_AMASS:AMASS_XT60-F_1x02_P7.20mm_Vertical | 5 | RS-Pro | 1.31 |
| XT60PW-M:AMASS_XT60PW-M | 3 | RS-Pro | 1.61 |
| Connector_IDC:IDC-Header_2x05_P2.54mm_Vertical | 4 | TE Connectivity | 1.77 |
| Connector_IDC:IDC-Header_2x06_P2.54mm_Vertical | 2 | Amphenol ICC | 1.50 |
| Connector_JST:JST_XH_B2B-XH-A_1x02_P2.50mm_Vertical | 7 | JST | 0.80 |
| Connector_JST:JST_XH_B2B-XH-A_1x02_P2.50mm_Vertical | 3 | JST | 0.80 |
| Connector_JST:JST_XH_B2B-XH-A_1x02_P2.50mm_Vertical | 6 | JST | 0.80 |
| Connector_JST:JST_XH_B3B-XH-AM_1x03_P2.50mm_Vertical | 3 | JST | 1.05 |
| Connector_JST:JST_XH_B3B-XH-AM_1x03_P2.50mm_Vertical | 1 | JST | 1.05 |
| Connector_JST:JST_XH_B3B-XH-AM_1x03_P2.50mm_Vertical | 1 | JST | 1.05 |
| Connector_JST:JST_XH_B4B-XH-A_1x04_P2.50mm_Vertical | 1 | JST | 1.19 |
| Connector_JST:JST_XH_B4B-XH-A_1x04_P2.50mm_Vertical | 4 | JST | 1.19 |
| Connector_JST:JST_XH_B4B-XH-A_1x04_P2.50mm_Vertical | 1 | JST | 1.19 |
| Connector_JST:JST_XH_B5B-XH-A_1x05_P2.50mm_Vertical | 4 | JST | 1.46 |
| Connector_JST:JST_XH_B6B-XH-A_1x06_P2.50mm_Vertical | 4 | JST | 2.65 |
| Connector_JST:JST_XH_B6B-XH-A_1x06_P2.50mm_Vertical | 4 | JST | 0.00 |
| Connector_JST:JST_XH_B10B-XH-AM_1x10_P2.50mm_Vertical | 1 | JST | 2.38 |
| Connector_PinHeader_2.54mm:PinHeader_1x02_P2.54mm_Vertical | 1 | RS-Pro | 2.07 |
| Connector_PinHeader_2.54mm:PinHeader_1x03_P2.54mm_Vertical | 20 | RS-Pro | 2.07 |

Fig 5. R4 BOM (excluding PCB)

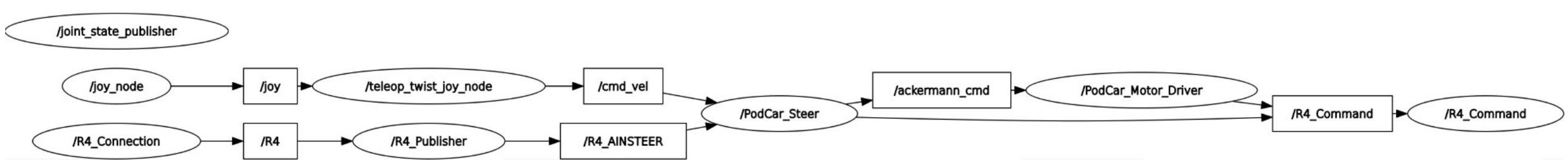

Fig 6. ROS2 interface node graph

## Hardware documentation / build instructions / files location

https://gitlab.com/charles.fox/r4pcb
BoM:  R4_hardware/R4_BOM.ods
Build instructions: R4_hardware/build_instructions.pdf
User guide: R4_UserGuide.pdf

# Discussion

R4 is designed to enable “no firmware programming” for end users, allowing them to immediately connect a selection of standard devices and program them using ROS2.  This allows, for example, a research robotics lab to be stocked with a set of standard components such as OSMC motor drivers, relays, and motors, which along with R4, a box of aluminium profile, and batteries could be used somewhat like a Lego collection to create many different robot forms. These could include mobile robots, robot arms, conveyor systems, and more. There is no current agreed definition of such a standard component set, but components with R4 support – either in this initial release or added later by the community – would define an informal standard. Together, R4 and this standard would then enable labs to stock with the same standard components and thus be able to rebuild one another’s designs and reproduce each others experiments with minimal ordering of additional components and minimal software reconfiguration. These users could include students as well as technicians and researchers, who are removed from the need to write firmware code and can need only in ROS2.

Firmware programmers who wish to add new device drivers to R4 are encouraged to do so and share them with end-users via the project. These are especially welcome for further open-source hardware devices.  It is hoped that a collection of such software drivers for R4 alongside OSH device hardware designs could grow to define a ecosystem or “distribution” of known, trusted, standard OSH components. Wrappers for other middleware such as YARP could be added around the text protocol by and for those who prefer them to ROS2. Forks of the physical PCB design are also welcome, such as creating new form factors optimised for different applications.

## Paper author contributions (CRediT)

Chris Waltham: Methodology, Investigation, Writing. Andy Perrett: Software. Rakshit Soni: ROS2 Software, Writing. Charles Fox: Methodology, Testing, Writing, Project administration.

## Acknowledgments

Lincoln MSc Robotics and Autonomous Systems students Abdul Basit, Maduabughichi Eze, Adeola Fagbenro, Offiong Okon Offiong, R.K.J. Charaka Prabath Ranathunga, Umit Karadeniz and Violet Mayne designed and built the OpenScout variant using R4. Fanta Camara and Kshitij Gaikwad developed previous OpenPodcar PCBs. Thanks to three anonymous reviewers for improving the paper.

## Funding statement

This work received Small Grant funding from the School of Computer Science, University of Lincoln.

## Competing interests

The authors declare that they have no competing interests.

## Licences

**R4 USER GUIDE**

Chris Waltham, Andy Perrett, Rakshit Soni, Charles Fox

## 1. Initial configuration

Do this once on obtaining an R4. (For example, having built one using the BoM and build instructions in R4_hardware/.)

Skill level: engineering technician.

### 1.1 R4 board acceptance tests

After building or obtaining an R4 board, ensure your Arduino Giga R1 is not connected to the R4 Board. Do not power your R4 board before completing the following tests.

First, check for shorts across the onboard 5V and 3.3V power regulators, the reference designators for these components are U1 and U2. The schematic describes the circuit in use and the PCB layout shows the physical positions of these regulators, refer to these documents and orientate yourself as to the locations of these parts. U1 and U2 are located near the 24V input XT60 connector on the short edge of the board. Make use of a multi meter in 'diode tests/continuity mode' to ensure there is no short between the output of the regulator and ground. Refer to the image above showing where these regulators are on the board and across which pins to make this test. *Ensure this is done prior to first powering the device.* Also ensure there is no short between the input of the regulator and the output of the regulator.

A short circuit across the regulators does not mean that the short is AT the regulator. This test can only tell you that a short exists somewhere on the 3.3V or 5V supply rails, should the test reveal a short, carefully inspect the surface mount IC's on the board for a solder bridge, this is the most likely place to find such a short. Continue visually checking until the source of all/any shorts are located and remove them with solder wick or other soldering techniques. The device will not work correctly and will almost certainly be damaged if this test is skipped. *Do not skip this test.*

Check every leg on the IC's to ensure that adjacent pins are not inadvertently connected together, clear any shorts found. Inspect the underside of the board (the through hole solder side), check every socket and through hole component for solder bridges. It is imperative that there are no inadvertent shorts or solder bridges on the board. *Take care and take your time.*

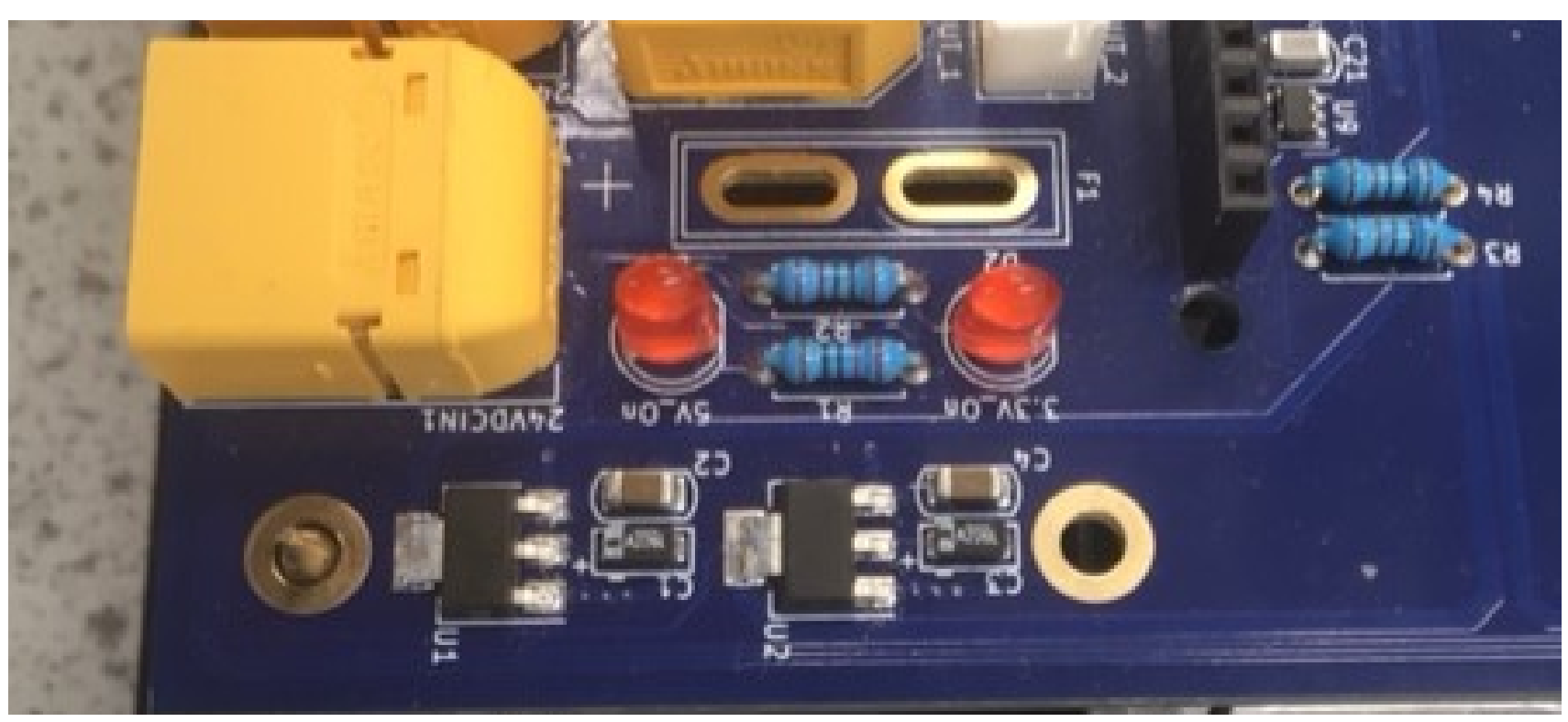

Figure 1: R4 Onboard Regulators Location

Check the I2C bus in the same way, ensure that the two I2C lines are not bridged anywhere.

These checks are the absolute minimum checks to perform, ideally a comparison between the schematic circuit and the boards real world connectivity should also be made. PCB board manufacturers can automate this check after assembly, although at significantly increased cost.

## 1.2 First Power On

U1 and U2 (onboard regulators for 5V and 3.3V) derive their input from the 12V XT60 input on the board. The first time the board is powered it is advised that an adjustable lab bench power supply with a current limiter is used to provide a 12V current limited supply to the 12V input port of R4. Set the lab power supply to 12V and limit the input current to approximately 100mA (do this in advance then switch off the lab PSU). This approach ensures that no more current than is required by R4 (when functioning correctly) can be delivered by the lab power supply. Make a power cable using the correct gender XT60 cable connector, the R4 PCB Silkscreen shows on the top layer of the board which connection is positive and which is negative. *Do not reverse polarity connect the board.* Double check that the connections made to the board are the correct way around, both at the R4 XT60 and at the external PSU. *Take your time, double check.*

Having fully checked your connections, ensure your power supply is set to 12V 100mA and switch on. Two LEDs (D1 and D2) will light if the regulators are working correctly, if these are not lit there is a fault, switch off quickly and re-do your initial checks.

The R4 board should have two lit LED's showing that the 3.3V and 5V supplies are present and the lab power supply will be showing 12V and about 100mA output. Using a multi-meter, set on a DC voltage range (the lowest range greater than 5V) carefully check without shorting any pins that output voltage at each regulator is as expected (i.e. 3.3V and 5V respectively). See Fig. 1.

One by one, check each IC has the correct voltage at its power pins, refer to the schematic and PCB file to see which supply should be present at each IC. Ensure these supplies are present and that no component on the board is excessively hot.

## 1.3 Attaching the Arduino Giga

Once it is is established that the R4 board is powered and the onboard voltage regulators output the correct voltages, checking for excessively hot components works well at this stage. Use of a thermal camera provides an easy safe method to do this. If this is not available, briefly touch the IC's on the board in turn to check for excessive heat, being very cautious as IC's can get very hot if not functioning correctly. It is unlikely that an IC will be extremely hot if the power supplies from the regulators are at the correct voltages. *Be cautious if you choose to test components for excessive heat by touch.*

Carefully engage all the pins on the R4 with the headers on the Arduino. This can be difficult and great care should be taken to ensure all the pins are engaged correctly. Do not press to hard at first, nudge pins such that they align correctly and coax the board onto the pins in stages. *It is easy to bend the pins – be careful.*

Check you can connect to the Arduino Giga via serial, the Giga will need to be programmed to test functionality. This involves either installing the Arduino IDE and the Giga board package or installing Visual Studio Code. Several Libraries are also required, a list of this is at the bottom of the read me on the GitHub.

Basic functionality of the R4 board in conjunction with the Arduino Giga is best first tested using I2C bus scanning firmware, an example of this code is present on the Github. The sketch required is called `R4_I2C_Bus` ↪ `Scanner.ino`. Compiling and programming the Arduino Giga with this program will report all I2C devices found on the bus. The scanner should report the devices shown in Fig. 3. Alternatively the screenshot shown in Fig. 3 is output when R4 first starts when the full firmware is installed.

## 1.4 Firmware configuration and installation

The firmware compiles in either Arduino Studio or Visual Studio Code. With the `R4_OSS_Software/R4OSSSoftware` ↪ `.ino` file selected in the IDE verify that the code compiles. R4's WiFi Communications uses UDP for the underlying protocol, it uses UDP because it is important that packets arrive at the receiver in the order sent. If TCP/IP were used ordered packet reception would not be guaranteed. WiFi communications have been tested using various routers and phone hot-spots. R4 connects using the typical WiFi SSID and password and WPA, WPA2 security. R4 Firmware currently needs to be hand coded to set WiFi parameters although later this will become part of a configuration setup function that will allow these parameters to be changed without having to modify the firmware. Two files require changes to enable use on different WiFi Networks:

*R4_UDP.h* : near the top this file, change the `SSID` and `SSID_PW` to those specific to your router or hot-spot:

```
R4 Test!

Heartbeat Frequency Hz : 2
TimedheartBeatlimiter  : 50
HeartbeatInterval_ms   : 500
heartbeatIntervalMax   : 1000

Datagram  Frequency Hz : 100
datagramInteruptlimiter: 10
datagramInterval_ms    : 10

Scanning i2c Bus...
i2c device found at address 0x20
i2c device found at address 0x40
i2c device found at address 0x41
i2c device found at address 0x42
i2c device found at address 0x61
i2c device found at address 0x70
Scan Complete
[TISR] Timer Input Freq (Hz) = 240000000
[TISR] Frequency =1000.00, _count = 1000
Started ITimer0 OK

Initializing Giga WiFi UDP Client
192.168.0.162
Connected to Wifi: R4-ROS2 : WL_CONNECTED
Starting Wifi UDP - 0 means failed: 1

Attempting to Establish a Connection: H:R4-ROS2;PNum:0;T:0;^0xb10b3a06^
Attempting to Establish a Connection: H:R4-ROS2;PNum:0;T:0;^0xb10b3a06^
Attempting to Establish a Connection: H:R4-ROS2;PNum:0;T:0;^0xb10b3a06^
Attempting to Establish a Connection: H:R4-ROS2;PNum:0;T:0;^0xb10b3a06^
Attempting to Establish a Connection: H:R4-ROS2;PNum:0;T:0;^0xb10b3a06^
```

Figure 2: R4 Initialisation Output

```
#define SSID "MyNetworkName"
#define SSID_PW "MyNetworkPassword"
```

*R4_UDP.cpp* : locate this line, and change the IP address to that of the host machine (PC, Pi, Laptop etc):

```
IPAddress udpServer = "192.168.0.117"; // your host computer
```

To enable analog sensor inputs, the firmware requires a simple change to the `R4_enableHardware()` function. Boolean parameters are present within this function to enable or disable specific hardware devices available on R4.

R4 is set to default 100Hz for data packets and 2Hz for heart beats with a 1 second timeout, these can be modified in the firmware though the defaults suffice for most applications.

After making the firmware changes, compile again and upload the compiled code to the Arduino Giga via a USB-C cable. Ensure the Giga is the selected board and that the COM port is properly detected and correctly set.

## 2. End-user instructions

The following are instructions for end-users. They assume that a build has been performed using the build instructions and the initial configuration completed using the above instructions.

Skill level: school or undergraduate engineering student.

### 2.1 Connecting power

Fig. 1 shows how the XT60's are wired at the power section of the PCB. The primary power input is marked '24VDCIN1' this connector is fused to the board via F1, this fuse can be selected theoretically up to a maximum of 20 amps depending on the application. This 24V supply is then distributed to two further XT60 connectors which allow external regulators to be added if application needs them. The design allows suitable external higher current 12V and 5V regulators to be driven by the 24V supply and returned to the board at the 12V '12IN1' and 5V (5IN1) input connectors, these external regulators can be specified by the user depending on the application. Recommendations for these are made in the documentation for the specific use cases examined so far. The supplies are then distributed around the board as needed, for example to the servo power pins or the H-bridge 12V connections. Refer to Fig. 1 for details of how these are wired. The power sub-systems are fused both at their output (24V) and at there inputs (12V and 5V respectively), allowing the user to specify current limits for these supplies by selecting appropriate external regulators and fuse ratings. It is also possible to use the 12V input independently – the R4 board does not require a 24V supply.

### 2.2 Connecting actuators

To connect up to two DHB12 motor drivers, R4 has two sockets marked DHB12-1 and DHB12-2, these have standard 12-pin IDC headers which connect via ribbon cables to the DHB12 H-bridges. When connecting these, ensure that pin one is aligned at the DHB12 H-bridge end as these are not supplied with notched headers. Failure to do this may result in damage to the R4, the DHB12's or both.

To connect to OSMC motor drivers, R4 has four sockets OSMC1 to OSMC4 with standard 10-pin IDC headers for ribbon cables to the drivers.

Servos can be plugged directly into the servo headers. These use single row 0.1" (2.54mm) pitch header sockets as per supplied on standard servos. If servos are used, ensure an external 5V regulator is connected to R4's 5V input connector. Ensure the servo cables' positions for 0V, Signal and 5V are aligned with those shown on the silk on the board.

Fig. 4 shows an example of DHB12 and motor connections to R4.

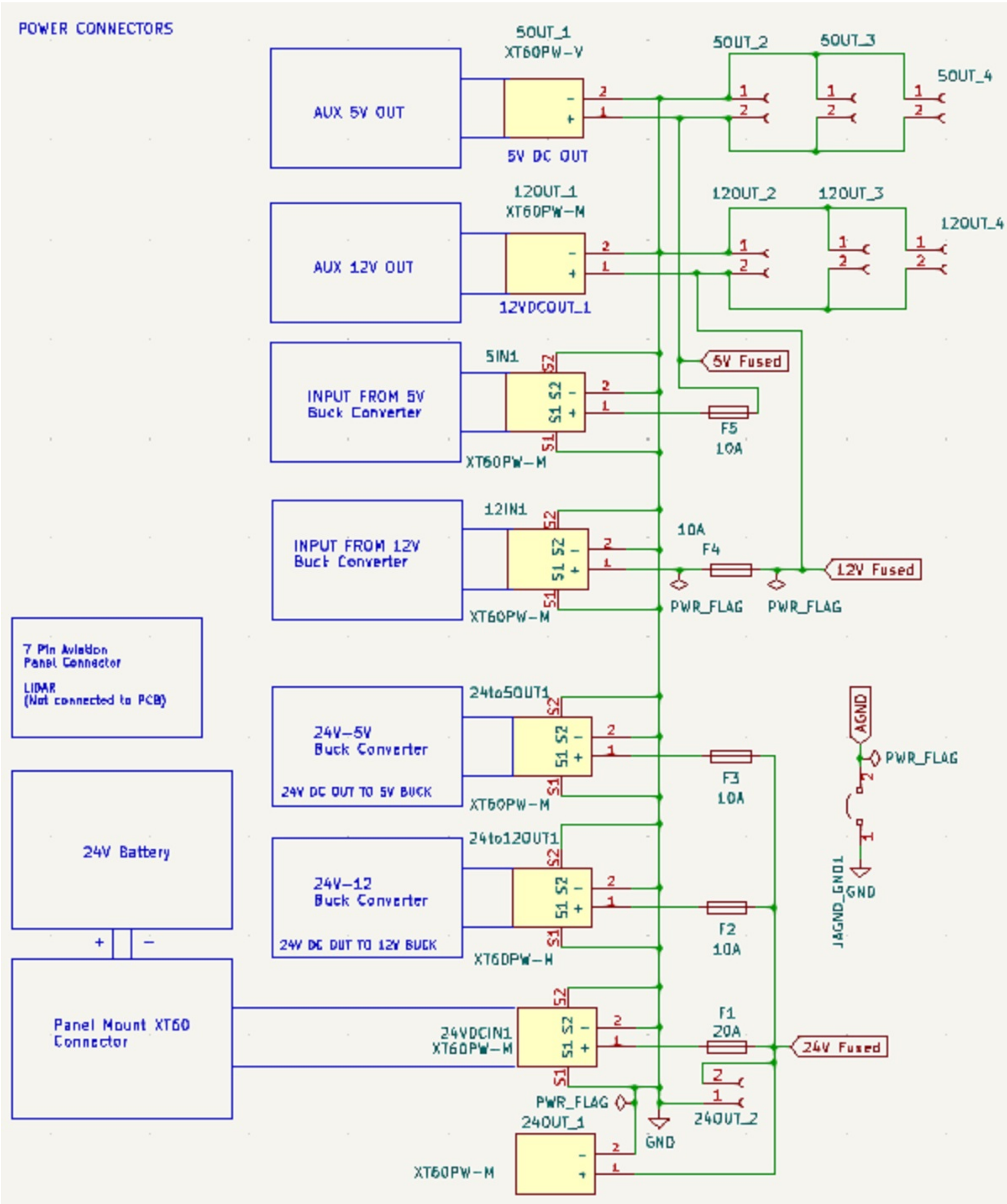

Figure 3: Power extract from the R4 Schematic

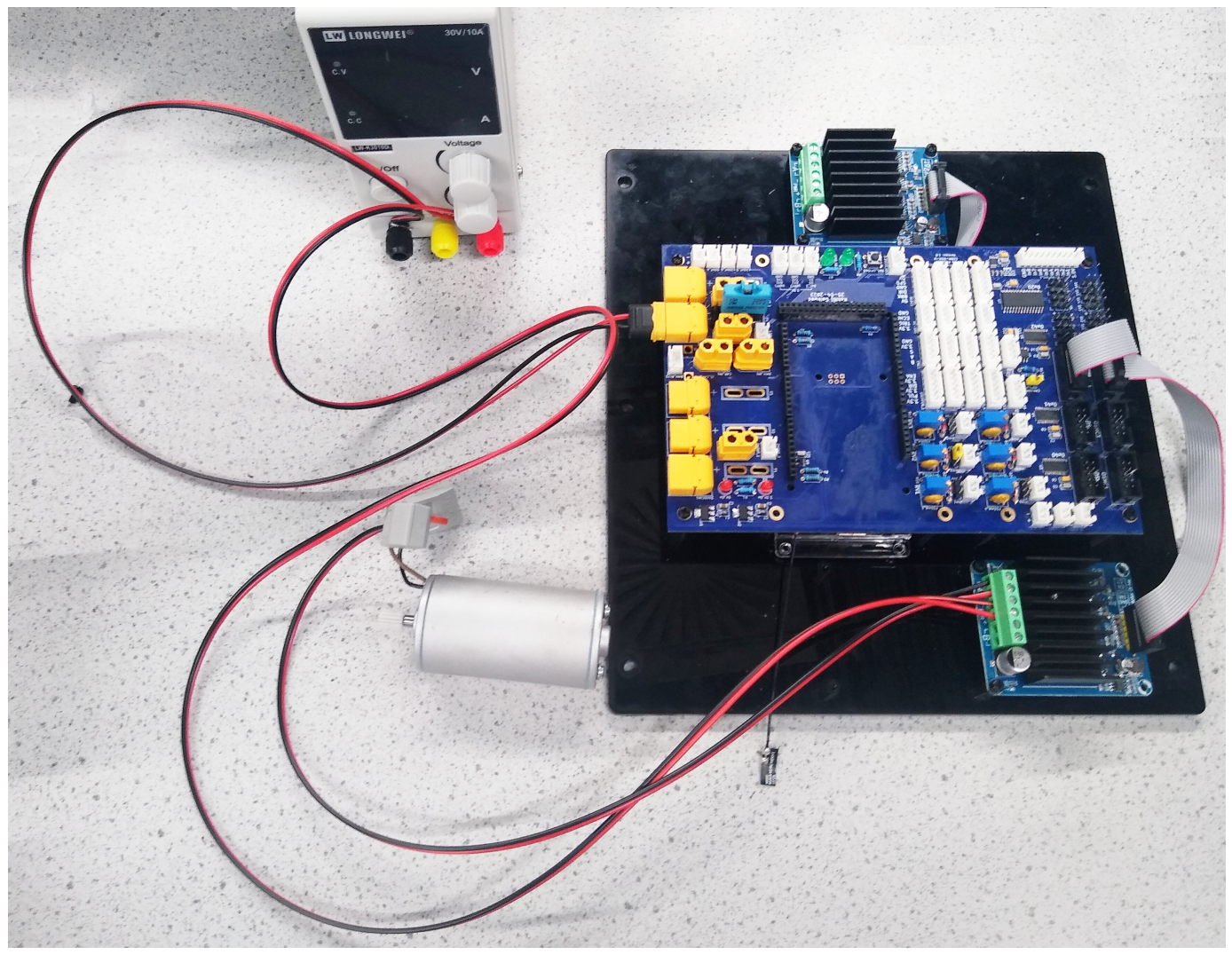

Figure 4: R4 bench setup with power supply and motor

### 2.3 Connecting sensors

To monitor up to 6 analog voltages, connect them to R4's analog inputs AIN1 to AIN6 using JST XH B2B 1x02P 2.50mm vertical connectors. The voltages on the analog input with be sent via the datagram (assuming the channel has been enabled in the firmware setup).

### 2.4 Adjusting ADC Voltage Ranges

The voltage range of the 16 bit ADC input channels provided by R4 are individually adjustable on the board. A Bourns 10 turn potentiometer is used to adjust, allowing the ADC input to span a specific voltage range. If the input is required to measure 0-12V for example the potentiometer can be adjusted such that it divides the incoming voltage by a ratio matching the input voltage range. This ensures that the range of values output by the ADC are equally distributed over the desired voltage range. Providing higher precision voltage measurements over the range required. 0-12V may be represented by 0 to 65535 at the ADC output alternatively, adjusting the potentiometer allows a range of 0 to 3V may to be represented by 0 to 65535 at the ADC output.

### 2.5 Using R4 via WiFi UDP Socket

R4's WiFi communications uses UDP for the underlying protocol, because it is important that packets arrive at the receiver in the order sent. If TCP was used, ordered packet reception could not be guaranteed.

The following is then a client program in Python, to run on a host PC, which sends motor and servo commands and reads status from the R4 board. Note that the client's obligations include checking and resending heartbeats, and adding times, packet numbers, and CRC checksums to messages.

```
import sys, zlib, time, errno, socket

def sendMsg(message, pNumSent, client_socket):
    message = message + ";T:" + str(int(time.perf_counter()))
    message = message + ";PNum:"+str(pNumSent)
    pNumSent += 1
    message = message + ";^"
    checksum = hex(zlib.crc32(message.encode()))
    message = message+checksum+"^"
    print("send: "+message)
    message=message.encode() #convert to python bytes
    client_socket.sendto(message, ("192.168.0.125", 2018)) #R4 ip,port
```

```
    return pNumSent


server_socket = socket.socket(socket.AF_INET, socket.SOCK_DGRAM)
server_socket.bind(("192.168.0.158", 2390)) # my own host ip, port
server_socket.setblocking(0)
client_socket = socket.socket(socket.AF_INET, socket.SOCK_DGRAM)
pNumSent=0

while True:
    #Read socket
    try:
        data = server_socket.recv(256) #buffer size
    except socket.error as e:
        if e.errno == errno.EWOULDBLOCK:
            continue #loop until we get something
        else:
            print(f"Error with UDP connection: {e}")
            break
    print("recv: "+data.decode())

    #Validate checksum
    data = data.split(b"^")
    checksum = data[1]
    data = data[0]+b"^"
    assert hex(zlib.crc32(data)) == checksum.decode()
    #Check if its a heartbeat, and send it back if so
    #(R4 will safety shut down if this is not done)
    firstKey = next(iter(data)) # heartbeat should always be the first param
    if firstKey == 72: #ascii H
        pNumSent = sendMsg("H:ROS2-R4",pNumSent,client_socket)
        lastHeartbeatTimestamp = time.perf_counter()
    #send some motor commands
    #dhb12, speed (PWM ms), dir, channel
    pNumSent = sendMsg("D:500,1,1", pNumSent, client_socket)
    pNumSent = sendMsg("D:500,0,2", pNumSent, client_socket)
    pNumSent = sendMsg("D:1000,1,3", pNumSent, client_socket)
    pNumSent = sendMsg("D:1000,0,4", pNumSent, client_socket)
```

### 2.6 Serial Port debug information

Debug information about R4 is output to the serial terminal. The Baud rate of this port is 1000000bps.

### 2.7 R4 Communications Protocol

R4 communicates with an external host computer using a human-readable ASCII line-based protocol. The protocol consists of carriage return terminated delimited strings (datagrams). The datagrams can be communicated either over a serial port or a UDP socket. R4 handles sampling hardware status, timing of datagrams, heartbeats, and asynchronous reception of command packets, this functionality is driven by Giga R1 hardware interrupts to maintain controlled timed sampling of the current hardware status.

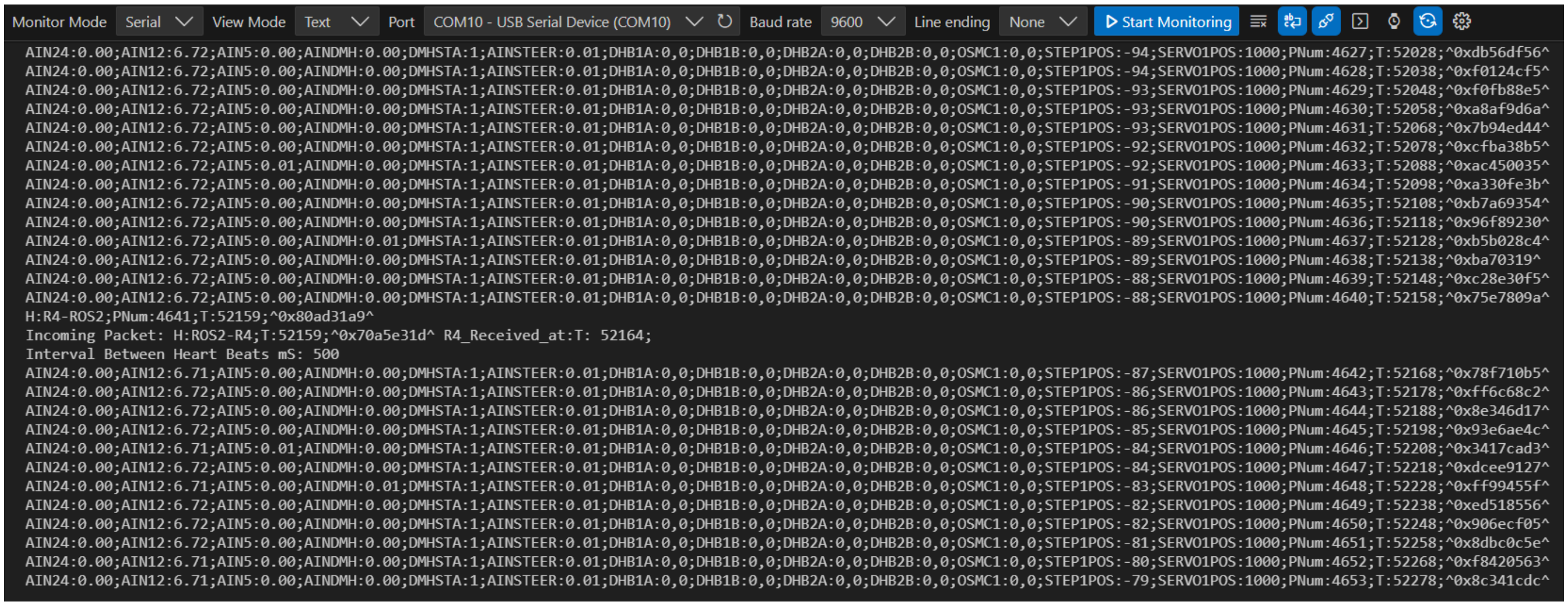

Figure 5: R4 Debug UDP Data via Serial Port

All datagrams are structured as: FieldName ':' (colon ends the Fieldname) DataElement ';' (semi colon denotes end of field). The simplest example of this is a FieldName with only one DataElement, such as

`'AIN24:24.18;'`

This has a FieldName 'AIN24' and a DataElement '24.18' the FieldName is separated from the DataElement by a ':' and the field is terminated with a ';'. Where a FieldName has several DataElements as per this case: 'O:400,1,1;' DataElements are separated by commas ','. This case specifies a PWM width and the direction to set the OSMC H-bridge and the OSMC channel (R4 has four separate OSMC channels).

### 2.7.1 Status datagrams

Datagrams are streamed out continuously by R4 during normal operation, reporting the state of the hardware at timed intervals, and consisting of strings in this format:

```
AIN24:24.18;AIN12:12.18;AIN5:5.13;AINDMH:12.15;DMHSTA:1;AINSTEER:1.06;DHB1A:4095,1;
OSMC1:400,1;STEP1POS:-100;SERVO1POS:2000;Pnum:7433;T:79080;^0xe164c64a^
```

The fields here mean the following:

```
AINxx: ADC voltages captured by R4
AINDMH: goes low when the DMH is pressed
DMHSTA: state of the DMH button (active low)
DHB1A :state of a DHB12 H Bridge Format is %On,Direction
OSMC1: state of an OSMC H Bridge Format is %On,Direction
STEP1POS  current position of stepper motor 1
SERVO1POS: position (microseconds PWM width) this represents the current angle of the servo
```

At the end of the datagram above are three further fields, 'PNum', 'T' and a 32-bit hex value enclosed by the caret character. 'PNum' is used to ensure packets are sequentially received, should the previously received packet number be greater than one different from the current packet number a packet has been missed. 'T' is used to determine how much time has elapsed since the previous packet, taking the difference between the previous packet time stamp and the current packet timestamp defines the periodicity of the datagram. The hex value is a 32-bit CRC checksum which may be used to detect data corruption across the link.

### 2.7.2 Command datagrams

Commands are sent to R4 using datagrams such as,

```
O:358,1,1;PNum:66;T:1699276044383;^0x7fefae81^
```

Breaking down the command packet (note the same field delimiters are used as for the datagram):

O:358,1,1,1; - The first character of the command packet received by R4 denotes the hardware sub component to be controlled. In this case ‘O’ means an OSMC H Bridge (note a ‘:’ ends the command field)

‘358,1,1’ instructs R4 to set the OSMC PWM state to 358 the first DataElement is ‘358’ (this controls the amount of time the H bridge is on per PWM cycle) a ‘,’ denotes the end of the first command DataElement. The following ‘1’ determines in which direction to set the OSMC H bridge (forwards or backwards) again a ‘,’ denotes the end of this DataElement. The last ‘1’ instructs the R4 board to set OSMC channel 1 to the preceding values. R4 board has four OSMC channels therefore we need to include a field which describes which channel to set.

Using this guidance, a user may receive data from R4 and make sense of the DataElements, the same method of FieldNaming, Delimeters and DataElement separation is used throughout. The command packets also use the same methods, each possible command is described in the R4 code with examples for usage. A table is also provided below for all currently implemented commands.

```
case ’H’: // HEARTBEAT Signal from UDP Server
// FORMAT = H: String "ROS2-R4"; PNum: integer xxx; integer T:xxxxx;uint32_t CRC32^
// EXAMPLE: H:ROS2-R4;PNum:1;T:xxxxx;^0xabcd1234^

case ’S’: // Set SERVO duration, channel
// FORMAT = S:int duration in milliseconds, int channel;uint32_t CRC32^
// EXAMPLE: S:560,1;PNum:1;T:xxxxx;^0xabcd1234^

case ’D’:// Set DHB12 speed and direction
// FORMAT = D:int duration in milliseconds,bool direction, int channel;uint32_t CRC32^
// EXAMPLE: D:560,1,1 to 4;PNum:1;T:xxxxx;^0xabcd1234^

case ’O’: // Set OSMC speed and direction
// FORMAT = O:int duration in milliseconds,bool direction, int channel;uint32_t CRC32^
// EXAMPLE: O:560,1,1;PNum:1;T:xxxxx;^0xabcd1234^

case ’R’: // Set DMH relay state
// FORMAT = R:int channel,bool state;PNum:xxx;T:xxxxx;uint32_t ^CRC32^
// EXAMPLE: R:1,On;PNum:1;T:1245;^0xabcd1234^

case ’E’: // emergency instruction to safely shutdown all devices
// FORMAT = E:String "STOP";uint32_t CRC32^
// EXAMPLE: E:STOP;PNum:1;T:xxxxx;^0xabcd1234^
```

New drivers may extend the list by adding their own relevant commands.

This is a plain ASCII implementation for clarity (over efficiency). The protocol may be easily extended with more datagram fields and command types. R4 firmware currently generates the hardware state datagram in the function below,

```
bool R4_makeUDPPacket(podcar_data_type data, bool restartPNum)
```

This function can be customised to generate any datagram required. Currently a set of enable states control which data elements are transmitted. The intention here is to standardise this datagram generation such that all ‘specific applications’ use the same standard hardware state datagram, elements of which are enabled and disabled during a protocol preamble as required. Standardisation of the state datagram here will mean that the R4 firmware will not need to be altered to make use of the board under all its use cases. R4’s datagram will simply report the state of all the ‘used’ hardware interfaces in every datagram transmitted. Likewise a standard set of commands packets for R4 covering all hardware interface settings requirements will be available for users, again without having to alter firmware, a ‘user’ simply makes use of only those state datagram fields and command packets needed for their particular application or use case. In order to minimise UDP traffic a preamble is intended which enables or disables specific hardware sub systems from the application side. This will mean only the ‘DataElements’ used by the application will be transmitted in the datagram.

### 2.7.3 Heartbeats

R4 periodically sends a heartbeat packet over the UDP WiFi link and the host on reception returns a host heartbeat packet. This has the specific function of determining link quality and does not change the state of the hardware under normal operation. R4 sends a heartbeat at half second intervals and if it does not receive a heartbeat within 1 second a ‘safe state’ is automatically established, this rate is ‘programmable’, should the host fail to return a heartbeat packet within a specified time frame R4’s hardware is placed in a safe state, motors, H-bridge output, Servos etc are stopped for safety reasons. The heartbeat is used to ensure the WiFi UDP link is communicating and stable. Where a ‘safe state’ has been initiated R4 switches modes automatically and repeatedly sends only heartbeat packets until the host responds with its response heartbeat. R4 attempts to reestablish the connection in this way and assumes on receipt of the hosts heartbeat that the link is now reestablished. At this time R4 revokes its ‘safe state’ mode and returns to normal operations, reestablishing a hardware state identical to that last set by the host. The heartbeat recovery is detectable at the serial terminal and at the hosts receiver by observing datagram transmissions restarting along with timed heartbeats. The link then continues as ‘normal’, timing heartbeats and reporting the state of the hardware via the state datagram messages.

R4 also monitors its own connection to the network. If the WiFi Connection is detected as lost then R4 also enters a safe state, disabling its hardware outputs. At this point it is currently necessary to RESET the micro controller, on reboot R4 attempts to reconnect to the WiFi Network. All R4 interfaces are held in a safe state until such time as a connection is reestablished. This prevents unsafe hardware states that could be caused by wireless link failures such as a robot going out of range or experiencing a noisy RF environment.

### 2.7.4 Example R4 communications stream capture

```
AIN24:24.28;AIN12:12.12;AIN5:5.12;AINDMH:12.11;DMHSTA:1;AINSTEER:1.03;DHB1A:4095,1;OSMC1:400,1;STEP1POS:-100;SERVO1POS:2000;PNum:7434;T:79090;^0xf07cfd9a^
AIN24:24.28;AIN12:12.18;AIN5:5.13;AINDMH:12.15;DMHSTA:1;AINSTEER:1.05;DHB1A:4095,1;OSMC1:400,1;STEP1POS:-100;SERVO1POS:2000;PNum:7435;T:79100;^0xfbd15ccb^
AIN24:24.21;AIN12:12.15;AIN5:5.11;AINDMH:12.19;DMHSTA:1;AINSTEER:1.03;DHB1A:4095,1;OSMC1:400,1;STEP1POS:-100;SERVO1POS:2000;PNum:7436;T:79110;^0xc86eb5d8^
AIN24:24.28;AIN12:12.11;AIN5:5.13;AINDMH:12.14;DMHSTA:1;AINSTEER:0.99;DHB1A:4095,1;OSMC1:400,1;STEP1POS:-100;SERVO1POS:2000;PNum:7437;T:79120;^0x97a9b72d^
AIN24:24.28;AIN12:12.19;AIN5:5.10;AINDMH:12.19;DMHSTA:1;AINSTEER:1.05;DHB1A:4095,1;OSMC1:400,1;STEP1POS:-100;SERVO1POS:2000;PNum:7438;T:79130;^0x3e19c07e^
AIN24:24.28;AIN12:12.16;AIN5:5.11;AINDMH:12.19;DMHSTA:1;AINSTEER:1.03;DHB1A:4095,1;OSMC1:400,1;STEP1POS:-100;SERVO1POS:2000;PNum:7439;T:79140;^0x71b76fdb^

Command ROS2 to R4
Incoming Packet: O:358,1,1;PNum:66;T:1699276044383;^0x7fefae81^ R4_Received_at:T: 79146;
(See the highlight:  the command setting above has been set by R4)
AIN24:24.28;AIN12:12.18;AIN5:5.08;AINDMH:11.83;DMHSTA:1;AINSTEER:1.02;DHB1A:4095,1;OSMC1:358,1;STEP1POS:-100;SERVO1POS:2000;PNum:7440;T:79150;^0x122e5cd7^
AIN24:24.28;AIN12:12.19;AIN5:5.08;AINDMH:12.29;DMHSTA:1;AINSTEER:1.00;DHB1A:4095,1;OSMC1:358,1;STEP1POS:-100;SERVO1POS:2000;PNum:7441;T:79160;^0x8415510f^
AIN24:24.31;AIN12:12.18;AIN5:5.12;AINDMH:12.12;DMHSTA:1;AINSTEER:1.05;DHB1A:4095,1;OSMC1:358,1;STEP1POS:-100;SERVO1POS:2000;PNum:7442;T:79170;^0xc98f4920^
AIN24:24.21;AIN12:12.15;AIN5:5.11;AINDMH:12.19;DMHSTA:1;AINSTEER:1.05;DHB1A:4095,1;OSMC1:358,1;STEP1POS:-100;SERVO1POS:2000;PNum:7443;T:79180;^0x76cae3fe^
AIN24:24.31;AIN12:12.15;AIN5:5.07;AINDMH:12.22;DMHSTA:1;AINSTEER:1.00;DHB1A:4095,1;OSMC1:358,1;STEP1POS:-100;SERVO1POS:2000;PNum:7444;T:79190;^0xcd5a71f8^
AIN24:24.21;AIN12:12.16;AIN5:5.11;AINDMH:12.19;DMHSTA:1;AINSTEER:1.00;DHB1A:4095,1;OSMC1:358,1;STEP1POS:-100;SERVO1POS:2000;PNum:7445;T:79200;^0x49caadf0^
H:R4-ROS2;PNum:7446;T:79201;^0x369e8e0a^
AIN24:24.28;AIN12:12.21;AIN5:5.07;AINDMH:12.19;DMHSTA:1;AINSTEER:1.05;DHB1A:4095,1;OSMC1:358,1;STEP1POS:-100;SERVO1POS:2000;PNum:7447;T:79210;^0x9e1296f0^
Incoming Packet: H:ROS2-R4;T:79201;^0xf390d9ad^ R4_Received_at:T: 79211;
Interval Between Heart Beats mS: 505
AIN24:24.21;AIN12:12.15;AIN5:5.09;AINDMH:12.15;DMHSTA:1;AINSTEER:0.99;DHB1A:4095,1;OSMC1:358,1;STEP1POS:-100;SERVO1POS:2000;PNum:7448;T:79220;^0xa00ea797^
AIN24:24.28;AIN12:12.15;AIN5:5.09;AINDMH:12.19;DMHSTA:1;AINSTEER:1.03;DHB1A:4095,1;OSMC1:358,1;STEP1POS:-100;SERVO1POS:2000;PNum:7449;T:79230;^0xcc20a001^
AIN24:24.28;AIN12:12.11;AIN5:5.11;AINDMH:12.15;DMHSTA:1;AINSTEER:0.99;DHB1A:4095,1;OSMC1:358,1;STEP1POS:-100;SERVO1POS:2000;PNum:7450;T:79240;^0xd9bd7e4^
Incoming Packet: O:1,0,1;PNum:67;T:1699276044483;^0xa13d0ac^ R4_Received_at:T: 79246;
AIN24:24.15;AIN12:12.16;AIN5:5.12;AINDMH:12.18;DMHSTA:1;AINSTEER:1.05;DHB1A:4095,1;OSMC1:1,0;STEP1POS:-100;SERVO1POS:2000;PNum:7451;T:79250;^0xa9f4eef8^
AIN24:24.28;AIN12:12.18;AIN5:5.11;AINDMH:12.09;DMHSTA:1;AINSTEER:1.00;DHB1A:4095,1;OSMC1:1,0;STEP1POS:-100;SERVO1POS:2000;PNum:7452;T:79260;^0xd019589c^
AIN24:24.18;AIN12:12.15;AIN5:5.11;AINDMH:12.15;DMHSTA:1;AINSTEER:0.86;DHB1A:4095,1;OSMC1:1,0;STEP1POS:-100;SERVO1POS:2000;PNum:7453;T:79270;^0xa2815607^
AIN24:24.18;AIN12:12.18;AIN5:5.12;AINDMH:12.19;DMHSTA:1;AINSTEER:0.99;DHB1A:4095,1;OSMC1:1,0;STEP1POS:-100;SERVO1POS:2000;PNum:7454;T:79280;^0xa440bdcd^
AIN24:24.33;AIN12:12.15;AIN5:5.12;AINDMH:12.15;DMHSTA:1;AINSTEER:1.00;DHB1A:4095,1;OSMC1:1,0;STEP1POS:-100;SERVO1POS:2000;PNum:7455;T:79290;^0x46fd08fa^
AIN24:24.28;AIN12:12.15;AIN5:5.09;AINDMH:12.22;DMHSTA:1;AINSTEER:1.00;DHB1A:4095,1;OSMC1:1,0;STEP1POS:-100;SERVO1POS:2000;PNum:7456;T:79300;^0xb47ff538^
AIN24:24.33;AIN12:12.16;AIN5:5.10;AINDMH:12.18;DMHSTA:1;AINSTEER:0.95;DHB1A:4095,1;OSMC1:1,0;STEP1POS:-100;SERVO1POS:2000;PNum:7457;T:79310;^0xf39c0a1b^
AIN24:24.31;AIN12:12.18;AIN5:5.11;AINDMH:12.12;DMHSTA:1;AINSTEER:1.07;DHB1A:4095,1;OSMC1:1,0;STEP1POS:-100;SERVO1POS:2000;PNum:7458;T:79320;^0x154abf30^
AIN24:24.13;AIN12:12.16;AIN5:5.12;AINDMH:12.18;DMHSTA:1;AINSTEER:0.98;DHB1A:4095,1;OSMC1:1,0;STEP1POS:-100;SERVO1POS:2000;PNum:7459;T:79330;^0xed1920cb^
AIN24:24.18;AIN12:12.18;AIN5:5.12;AINDMH:12.16;DMHSTA:1;AINSTEER:1.00;DHB1A:4095,1;OSMC1:1,0;STEP1POS:-100;SERVO1POS:2000;PNum:7460;T:79340;^0x306eb6ac^
Incoming Packet: O:311,0,1;PNum:68;T:1699276044583;^0xf8720e0^ R4_Received_at:T: 79347;
AIN24:24.28;AIN12:12.16;AIN5:5.11;AINDMH:12.19;DMHSTA:1;AINSTEER:0.95;DHB1A:4095,1;OSMC1:311,0;STEP1POS:-100;SERVO1POS:2000;PNum:7461;T:79350;^0xb87164d4^
Incoming Packet: D:4095,0,1;PNum:69;T:1699276044587;^0x63f5e079^ R4_Received_at:T: 79351;
AIN24:24.13;AIN12:12.08;AIN5:5.07;AINDMH:11.93;DMHSTA:1;AINSTEER:0.86;DHB1A:4095,0;OSMC1:311,0;STEP1POS:-100;SERVO1POS:2000;PNum:7462;T:79360;^0x1a020faa^
AIN24:24.28;AIN12:12.15;AIN5:5.11;AINDMH:12.14;DMHSTA:1;AINSTEER:0.94;DHB1A:4095,0;OSMC1:311,0;STEP1POS:-100;SERVO1POS:2000;PNum:7463;T:79370;^0x145b511a^
AIN24:24.21;AIN12:12.15;AIN5:5.11;AINDMH:12.22;DMHSTA:1;AINSTEER:0.98;DHB1A:4095,0;OSMC1:311,0;STEP1POS:-100;SERVO1POS:2000;PNum:7464;T:79380;^0x3f03c38f^
AIN24:24.28;AIN12:12.15;AIN5:5.11;AINDMH:12.11;DMHSTA:1;AINSTEER:0.99;DHB1A:4095,0;OSMC1:311,0;STEP1POS:-100;SERVO1POS:2000;PNum:7465;T:79390;^0x3266daa7^
AIN24:24.31;AIN12:12.14;AIN5:5.11;AINDMH:12.16;DMHSTA:1;AINSTEER:0.95;DHB1A:4095,0;OSMC1:311,0;STEP1POS:-100;SERVO1POS:2000;PNum:7466;T:79400;^0x865722f2^
AIN24:24.28;AIN12:12.18;AIN5:5.11;AINDMH:12.21;DMHSTA:1;AINSTEER:0.92;DHB1A:4095,0;OSMC1:311,0;STEP1POS:-100;SERVO1POS:2000;PNum:7467;T:79410;^0x6a6804ba^
AIN24:24.18;AIN12:12.18;AIN5:5.08;AINDMH:12.18;DMHSTA:1;AINSTEER:0.98;DHB1A:4095,0;OSMC1:311,0;STEP1POS:-100;SERVO1POS:2000;PNum:7468;T:79420;^0x15deab73^
AIN24:24.28;AIN12:12.12;AIN5:5.11;AINDMH:12.19;DMHSTA:1;AINSTEER:0.99;DHB1A:4095,0;OSMC1:311,0;STEP1POS:-100;SERVO1POS:2000;PNum:7469;T:79430;^0xee602e95^
AIN24:24.28;AIN12:12.16;AIN5:5.09;AINDMH:12.16;DMHSTA:1;AINSTEER:0.94;DHB1A:4095,0;OSMC1:311,0;STEP1POS:-100;SERVO1POS:2000;PNum:7470;T:79440;^0xd24c81f4^
Incoming Packet: O:400,0,1;PNum:70;T:1699276044683;^0x5b016351^ R4_Received_at:T: 79446;
AIN24:24.28;AIN12:12.15;AIN5:5.12;AINDMH:12.18;DMHSTA:1;AINSTEER:1.00;DHB1A:4095,0;OSMC1:400,0;STEP1POS:-100;SERVO1POS:2000;PNum:7471;T:79450;^0xb1870b6e^
```

```
AIN24:24.18;AIN12:12.16;AIN5:5.15;AINDMH:12.08;DMHSTA:1;AINSTEER:0.98;DHB1A:4095,0;OSMC1:400,0;STEP1POS:-100;SERVO1POS:2000;PNum:7472;T:79460;^0xf6affe68^
AIN24:24.28;AIN12:12.18;AIN5:5.13;AINDMH:12.16;DMHSTA:1;AINSTEER:0.99;DHB1A:4095,0;OSMC1:400,0;STEP1POS:-100;SERVO1POS:2000;PNum:7473;T:79470;^0xda485031^
AIN24:24.13;AIN12:12.15;AIN5:5.11;AINDMH:12.16;DMHSTA:1;AINSTEER:0.99;DHB1A:4095,0;OSMC1:400,0;STEP1POS:-100;SERVO1POS:2000;PNum:7474;T:79480;^0xa3df267e^
AIN24:24.28;AIN12:12.15;AIN5:5.09;AINDMH:12.11;DMHSTA:1;AINSTEER:0.99;DHB1A:4095,0;OSMC1:400,0;STEP1POS:-100;SERVO1POS:2000;PNum:7475;T:79490;^0x57a7fa5f^
AIN24:24.33;AIN12:12.16;AIN5:5.09;AINDMH:12.18;DMHSTA:1;AINSTEER:1.00;DHB1A:4095,0;OSMC1:400,0;STEP1POS:-100;SERVO1POS:2000;PNum:7476;T:79500;^0x71e1d3c8^
AIN24:24.28;AIN12:12.11;AIN5:5.11;AINDMH:12.12;DMHSTA:1;AINSTEER:0.99;DHB1A:4095,0;OSMC1:400,0;STEP1POS:-100;SERVO1POS:2000;PNum:7477;T:79510;^0x1250a253^
AIN24:24.18;AIN12:12.09;AIN5:5.12;AINDMH:12.18;DMHSTA:1;AINSTEER:0.99;DHB1A:4095,0;OSMC1:400,0;STEP1POS:-100;SERVO1POS:2000;PNum:7478;T:79520;^0x8cac5a1d^
AIN24:24.28;AIN12:12.09;AIN5:5.13;AINDMH:12.14;DMHSTA:1;AINSTEER:1.19;DHB1A:4095,0;OSMC1:400,0;STEP1POS:-100;SERVO1POS:2000;PNum:7479;T:79530;^0x725f5219^
AIN24:24.56;AIN12:12.16;AIN5:5.11;AINDMH:12.18;DMHSTA:1;AINSTEER:0.98;DHB1A:4095,0;OSMC1:400,0;STEP1POS:-100;SERVO1POS:2000;PNum:7480;T:79540;^0x75743634^
AIN24:24.28;AIN12:12.08;AIN5:5.11;AINDMH:12.16;DMHSTA:1;AINSTEER:1.03;DHB1A:4095,0;OSMC1:400,0;STEP1POS:-100;SERVO1POS:2000;PNum:7481;T:79550;^0x72aa1f23^
AIN24:24.28;AIN12:12.15;AIN5:5.11;AINDMH:12.11;DMHSTA:1;AINSTEER:1.03;DHB1A:4095,0;OSMC1:400,0;STEP1POS:-100;SERVO1POS:2000;PNum:7482;T:79560;^0x400f7e78^
AIN24:24.31;AIN12:12.11;AIN5:5.11;AINDMH:12.12;DMHSTA:1;AINSTEER:0.92;DHB1A:4095,0;OSMC1:400,0;STEP1POS:-100;SERVO1POS:2000;PNum:7483;T:79570;^0xac5d508e^
AIN24:24.28;AIN12:12.18;AIN5:5.10;AINDMH:12.08;DMHSTA:1;AINSTEER:1.05;DHB1A:4095,0;OSMC1:400,0;STEP1POS:-100;SERVO1POS:2000;PNum:7484;T:79580;^0x266846aa^
AIN24:24.33;AIN12:12.15;AIN5:5.12;AINDMH:12.18;DMHSTA:1;AINSTEER:1.00;DHB1A:4095,0;OSMC1:400,0;STEP1POS:-100;SERVO1POS:2000;PNum:7485;T:79590;^0x3663b151^
AIN24:24.28;AIN12:12.15;AIN5:5.12;AINDMH:12.16;DMHSTA:1;AINSTEER:0.98;DHB1A:4095,0;OSMC1:400,0;STEP1POS:-100;SERVO1POS:2000;PNum:7486;T:79600;^0xb7f4e3c6^
AIN24:24.28;AIN12:12.15;AIN5:5.11;AINDMH:12.21;DMHSTA:1;AINSTEER:1.02;DHB1A:4095,0;OSMC1:400,0;STEP1POS:-100;SERVO1POS:2000;PNum:7487;T:79610;^0xd3af53c9^
AIN24:24.28;AIN12:12.08;AIN5:5.12;AINDMH:12.12;DMHSTA:1;AINSTEER:0.95;DHB1A:4095,0;OSMC1:400,0;STEP1POS:-100;SERVO1POS:2000;PNum:7488;T:79620;^0x7dc2e781^
AIN24:24.28;AIN12:12.21;AIN5:5.09;AINDMH:12.09;DMHSTA:1;AINSTEER:1.03;DHB1A:4095,0;OSMC1:400,0;STEP1POS:-100;SERVO1POS:2000;PNum:7489;T:79630;^0x387f458e^
AIN24:24.28;AIN12:12.19;AIN5:5.12;AINDMH:12.23;DMHSTA:1;AINSTEER:1.06;DHB1A:4095,0;OSMC1:400,0;STEP1POS:-100;SERVO1POS:2000;PNum:7490;T:79640;^0xab35b51a^
AIN24:24.31;AIN12:12.08;AIN5:5.09;AINDMH:12.16;DMHSTA:1;AINSTEER:1.03;DHB1A:4095,0;OSMC1:400,0;STEP1POS:-100;SERVO1POS:2000;PNum:7491;T:79650;^0xf80ca3ad^
AIN24:24.21;AIN12:12.12;AIN5:5.10;AINDMH:12.21;DMHSTA:1;AINSTEER:1.05;DHB1A:4095,0;OSMC1:400,0;STEP1POS:-100;SERVO1POS:2000;PNum:7492;T:79660;^0xb2e926b6^
AIN24:24.28;AIN12:12.18;AIN5:5.11;AINDMH:12.12;DMHSTA:1;AINSTEER:1.03;DHB1A:4095,0;OSMC1:400,0;STEP1POS:-100;SERVO1POS:2000;PNum:7493;T:79670;^0xb9feb3b5^
AIN24:24.28;AIN12:12.16;AIN5:5.11;AINDMH:12.16;DMHSTA:1;AINSTEER:1.05;DHB1A:4095,0;OSMC1:400,0;STEP1POS:-100;SERVO1POS:2000;PNum:7494;T:79680;^0xd17a81c7^
AIN24:24.31;AIN12:12.08;AIN5:5.09;AINDMH:12.15;DMHSTA:1;AINSTEER:0.98;DHB1A:4095,0;OSMC1:400,0;STEP1POS:-100;SERVO1POS:2000;PNum:7495;T:79690;^0x373f8009^
AIN24:24.13;AIN12:12.15;AIN5:5.12;AINDMH:12.14;DMHSTA:1;AINSTEER:0.99;DHB1A:4095,0;OSMC1:400,0;STEP1POS:-100;SERVO1POS:2000;PNum:7496;T:79700;^0x25af4383^
H:R4-ROS2;PNum:7497;T:79701;^0xa15fdf54^
Incoming Packet: H:ROS2-R4;T:79701;^0x3b7056dd^ R4_Received_at:T: 79705;
Interval Between Heart Beats mS: 495
```

### 2.8 Using R4 with ROS2

The ROS2 interface wraps the above UDP connection as ROS2 nodes, which automatically manage the required timestamps, CRC and heartbeat checks and enable the user to read and write R4 messages. Their functionalities are separated to simplify synchronization, by making use of ROS2 parallelism.

*R4_Connection.py* handles the low level UDP socket and heartbeat responses, this publishes one topic, `R4`, containing the entire UDP data. It also publishes two further topics to signal the status of the R4 board: `R4_Active` is set to True when the communications link to the board is working; and `R4_Heart_Beat` is the heartbeat, toggling between True and False. Should anything fail, any errors get detected, or a timeout value is reached during initialisation, the node will *reset* itself by closing the socket and keeping the topic *Active* false, and start up again. This allows for a different power up order between R4 and the ROS2 host node.

*R4_Publisher.py* subscribes to the `R4` topic, splits the UDP data fields, and publishes them as separate ROS2 topics. The new topic names are created as the R4 protocol field names, prefixed with 'R4_'.

*R4_Receiver.py* subscribes to `R4_Command` containing user R4 commands as strings of R4 protocol, and handles sending them out to the R4 board.

The R4 board's IP address should be edited in both *R4_Connenction.py* and *R4_Receiver.py* to match what is used.

After running all three nodes commands can be sent to individual R4 devices, such as this to set the speed of motor 1:

```
$ ros2 topic pub /R4_Command std_msgs/String "data: D:500,1,1" -1
```

and individual R4 channels can be inspected such as

```
$ ros2 topic echo R4_DBH1A
```

The nodes are in R4_ROS2HostInterface, and run on a host PC on ROS2. The ROS2 interface is tested with Ubuntu 22.04.

(If using ROS2 inside a container, then the host R4 port must be forwarded *as UDP* such as by `docker ↪ run -p 2390:2390/udp -it osrf/ros:humble-desktop` .)

The intended use case is that robotics projects can then create their own ROS2 nodes to represent the particular actuators and sensors as used their robots, and translate messages about what they want their robots to do into individual R4 commands to pass to and from the R4 ROS2 nodes. These will be project-specific, for example if a DHB12 is used to control a linear actuator used for pushing the Ackermann steering of a vehicle, its ROS2 node will need to take an application-appropriate input such as an angle or Twist message, and use Ackermann geometry to convert it to the required DHB12 values to pass via the serial protocol. (This particular example can be found at exampleUses/OpenPodCar_V2/pod2_bringup/ROS2_R4_nodes/Podcar_Steer_node.py; various other examples are provided in the validation test examples.)